\documentclass[conference]{IEEEtran}
\IEEEoverridecommandlockouts
\usepackage{cite}
\usepackage{amsthm}
\usepackage{amsmath,amssymb,amsfonts}
\usepackage{algorithmic}
\usepackage{graphicx}
\usepackage{subfigure}
\usepackage{textcomp}
\usepackage{xcolor}
\usepackage{color}
\usepackage{siunitx}
\def\BibTeX{{\rm B\kern-.05em{\sc i\kern-.025em b}\kern-.08em
    T\kern-.1667em\lower.7ex\hbox{E}\kern-.125emX}}

\begin{document}

\title{Potential Field-based Path Planning with Interactive Speed Optimization for Autonomous Vehicles


}

\author{Pengfei Lin$^{1}$, Ehsan Javanmardi$^{1}$, Jin Nakazato$^{1}$, and Manabu Tsukada$^{1}$
\thanks{$^{1}$P. Lin, E. Javanmardi, J. Nakazato, and M. Tsukada are with the Dept. of Creative Informatics, The University of Tokyo, Tokyo 113-8657, Japan. (e-mail: \{linpengfei0609, ejavanmardi, jin-nakazato, mtsukada\}@g.ecc.u-tokyo.ac.jp)}
}

\maketitle

\begin{abstract}
Path planning is critical for autonomous vehicles (AVs) to determine the optimal route while considering constraints and objectives. The potential field (PF) approach has become prevalent in path planning due to its simple structure and computational efficiency. However, current PF methods used in AVs focus solely on the path generation of the ego vehicle while assuming that the surrounding obstacle vehicles drive at a preset behavior without the PF-based path planner, which ignores the fact that the ego vehicle's PF could also impact the path generation of the obstacle vehicles. To tackle this problem, we propose a PF-based path planning approach where local paths are shared among ego and obstacle vehicles via vehicle-to-vehicle (V2V) communication. Then by integrating this shared local path into an objective function, a new optimization function called interactive speed optimization (ISO) is designed to allow driving safety and comfort for both ego and obstacle vehicles. The proposed method is evaluated using MATLAB/Simulink in the urgent merging scenarios by comparing it with conventional methods. The simulation results indicate that the proposed method can mitigate the impact of other AVs' PFs by slowing down in advance, effectively reducing the oscillations for both ego and obstacle AVs.
\end{abstract}


\section{Introduction}\label{intro}

Due to the surge in traffic accidents, autonomous vehicles (AVs) have captured public attention and are poised to revolutionize transportation. Many road tests have been conducted to promote the commercialization of AVs, indicating growing interest and potential. Nevertheless, since 2014, the Department of Motor Vehicles in California has received over 570 reports of collisions involving autonomous vehicles (AVs) \cite{AVreport2021, McCarthy2022-fp}, which has prompted concerns among the public regarding the safety of AVs currently on the road. Safe and efficient path-planning algorithms are crucial for ensuring the driving safety of AVs by searching for and identifying a safe and feasible path that avoids collisions with other road users.

Potential field (PF)-based path planning is becoming more popular for autonomous vehicles (AVs) due to its simple structure and real-time computational capabilities as a risk-aware method. Originally devised for solving planning problems in manipulators and mobile robots \cite{Khatib1986-dv}, the PF method involves creating an attractive potential field at the target point and repulsive potential fields around obstacles. PF-based path planning is gaining traction in the autonomous vehicle (AV) industry, as it offers a promising solution for navigating complex environments with a focus on safety. This approach has the potential to contribute to the evolution of AV path-planning strategies.
\begin{figure*}[t]
    \centering
    \includegraphics[width=0.95\hsize]{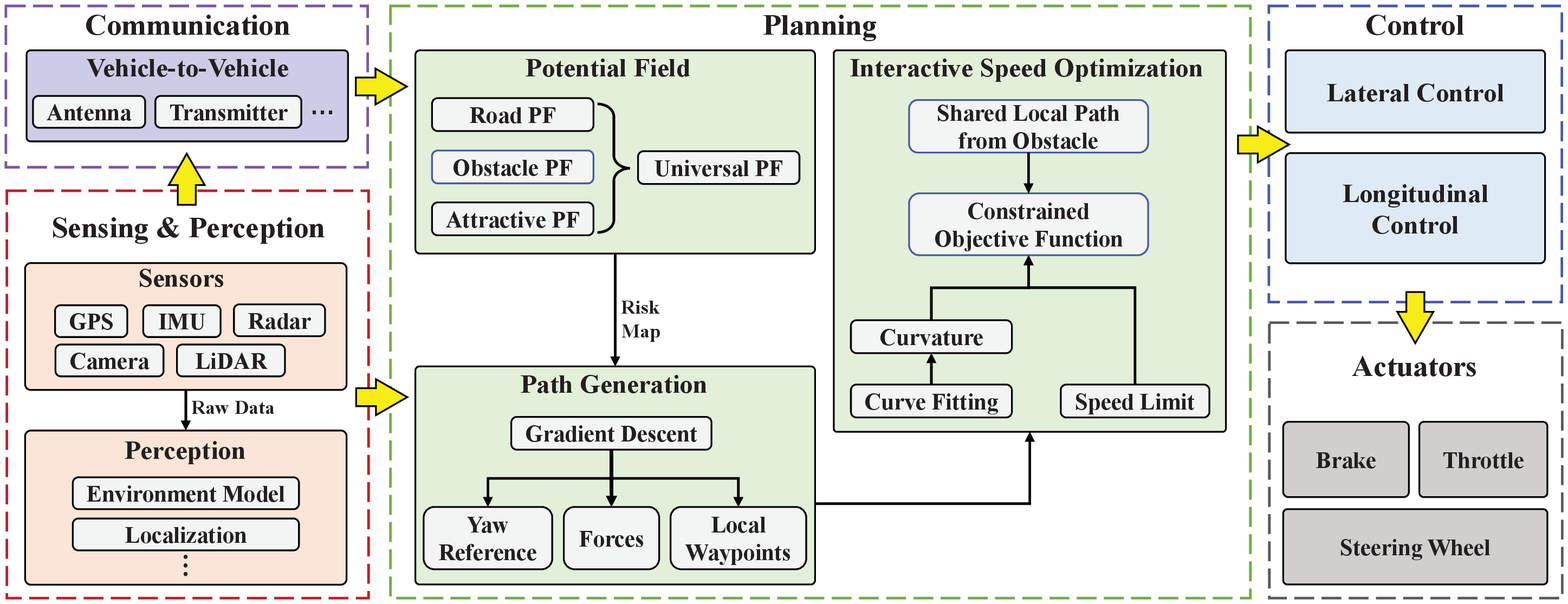}
    \caption{Overall system framework (Sensing \& Perception, Communication, Planning, Control, and Actuators): Planning layer extracts the shared local path from V2V communication and then formulates a constrained objective function with proper analysis}
    \label{system_scheme}
\end{figure*}

PF approach has been widely used to handle various driving scenarios, such as collision avoidance on highways and curved roads. However, its application in merging scenarios has received less attention than other road structures. Besides, the assumption that only the ego (host) vehicle has the PF-based path planner is commonly-acquiescent. Therefore, in the simulation stage, the obstacle vehicles are assumed to drive along a predetermined path while unilaterally focusing on the path generation of the ego vehicle. This paper proposes a unique interactive speed optimization for PF-based path planning among connected AVs equipped with V2V communication, as depicted in Fig. \ref{system_scheme}. The major contributions are concluded as follows.
\begin{itemize}
    \item Introduce a piece-wise road potential function for the merging road where two lanes merge into one lane. 
    \item Propose an interactive speed optimization (ISO) by solving a constrained objective function that considers driving safety, ride comfort, and the shared local path (SLP) from other connected AVs.
\end{itemize}
The structure of this paper is as follows: Section II reviews the related work on PF-based path planning. Section III describes the potential functions used for generating the PF. Section IV explains the proposed ISP method. Section V presents the comparative simulation results. Finally, Section VI concludes the paper. 

\section{Related Work}\label{related_work}

This section reviews research on PF-based path planning in autonomous driving. Early in 1995, Lopez et al. \cite{Lopez1995-pr} used the artificial potential function to guide maneuvering chase vehicles, which defines a suitable scalar function for the locality of the target vehicle with a set of bounded impulses by considering the obstructions and path constraints. Leonard et al. \cite{Leonard2001-ir} introduced a pioneering framework that utilizes artificial PF and virtual leaders to achieve coordinated and distributed control of multiple AVs, and a Lyapunov function is constructed to prove the closed-loop stability. However, the proposed method's effectiveness was only theoretically demonstrated, as no simulations or experimental tests were provided. Baronov et al. \cite{Baronov2008-zf} introduced a plan and control framework for AVs to ascend or descend along a PF to track a single target, but their simulation study did not incorporate collision avoidance. Wolf et al. \cite{Wolf2008-ye} used exponential and trigonometric functions to design the artificial potential functions for AVs, which built a triangle PF directly behind the AVs. Nevertheless, the proposed approach could not address the issue of local minima, as described in \cite{Koren1991-ar}, and the merging scenario was not discussed. Hesse and Sattel \cite{Hesse2007-ho} presented an extended approach to incorporate vehicle dynamics in an elastic band immersed in a PF-based hazard map; nevertheless, the simulation study only simulated a single static obstacle. Ji et al. \cite{Ji2017-fk} integrated the PF approach with a multi-constrained model predictive control (MMPC) technique to address path planning and tracking tasks for AVs. However, the generated path did not consistently adhere to vehicle dynamics, and their study only considered a single forward obstacle with constant speed or acceleration. Rasekhipour et al. Recently, Wang et al. \cite{Wang2019-nf, Wang2020-xl} introduced a novel approach incorporating crash severity and the PF into the objective function of a model predictive control (MPC) framework aimed at minimizing crash severity. However, the case studies assumed that collisions were unavoidable. Similarly, Lin et al. \cite{Lin2020-ry, Lin2020-uc} investigated the integration of potential fields (PF) with a clothoid curve for collision avoidance in a waypoint tracking scenario. However, their study made the assumption that the obstacles were moving solely on a straight road without any lane-changing behavior. Latest, Xie et al. \cite{Xie2022-jm} proposed a distributed motion planning framework utilizing the PF for overtaking scenarios. They designed a bounded distributed control protocol for multiple AVs. However, the merging scenario was not discussed, and the PF-based motion planner was only implemented for the lead AV. Wang et al. \cite{Wang2022-tt} combined a banana-shaped (ellipse-like) potential field and a linear time-varying (LTV) MPC to address overtaking maneuvers on large curvy roads, considering drivers’  handling characteristics. Still, The focus of the study was limited to low-speed obstacles with a single predefined motion. Similarly, Wu et al.~\cite{Wu2022-ae} proposed a PF-based lane-change algorithm for generating human-like trajectories while considering risks, drivers' focus shifts, and speed requirements. However, they did not consider the merging scenario and predefined the obstacle behaviors for observing the ego vehicle's path.
\vspace{0.5cm}

\section{Potential Field for Collision Avoidance}\label{pf}

In this section, we will explore the potential functions used to establish the PF, as well as the proposed cubic polynomial that is used to process the waypoints.
\begin{figure*}[t]
    \centering
    \includegraphics[width=0.95\hsize]{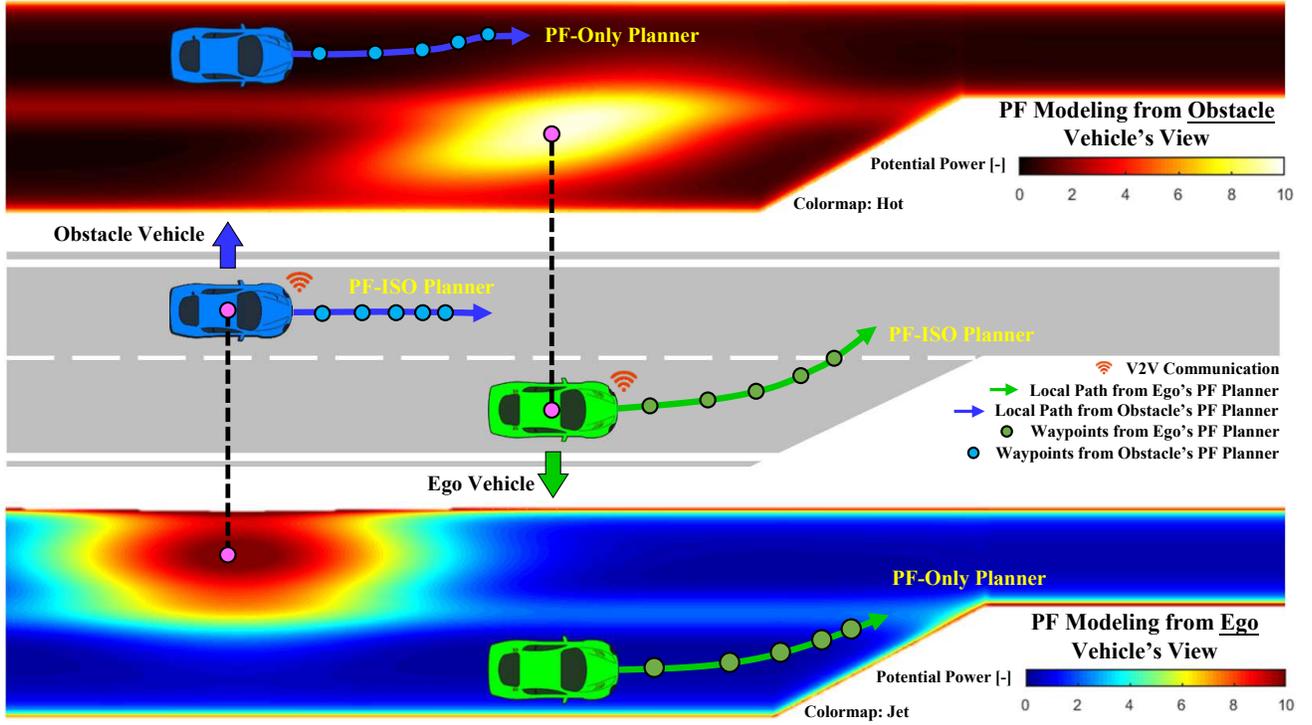}
    \caption{Merging scenario where the obstacle (blue) vehicle and the ego (green) vehicle apply the PF independently (Colormap: Hot for the blue vehicle and Colormap: Jet for the green vehicle). The paths depicted on the colormap are generated by the PF-only planner, while the paths denoted in the middle part are produced by the PF-ISO planner; the distance between the waypoints varies with speed}
    \label{inte_apf}
\end{figure*}

\subsection{Potential Functions}

When applying the PF algorithm to road structures, the potential functions are comprised of three distinct parts: the attractive PF, the road PF, and the obstacle PF.

\subsubsection{Attractive Potential Function} To propel the vehicle forward, the attractive potential function $U_{AT}$ is designed to generate a driving force.
\begin{equation}
    U_{AT}=
    \frac{1}{2}\lambda \left(X-X_{tar}\right)^2,
    \label{attr_pf}
\end{equation}
Here, $\lambda$ represents the slope scale, while $X$ and $X_{tar}$ refer to the longitudinal positions of the vehicle and the target point, respectively.

\subsubsection{Road Potential Function} The construction of a road typically involves three main components: road edges, lanes, and lane dividers. First, the potential function used for the lane divider is formulated below:
\begin{equation}
    U_{LD} =
    A_{lane}\exp{-\frac{(Y-Y_{lane})^2}{2\sigma^2}},
    \label{pf_lane}
\end{equation}
where $A_{lane}$ is the amplitude of the lane divider's PF, and $Y$ denotes the lateral position of the ego vehicle. $Y_{lane}$ is the lateral position of the lane divider. $\sigma$ refers to the rising/falling slope of the lane potential. To accommodate merge roads, we adapted the potential function for the road edges using the following expression.
\newpage
\begin{equation}
    U_{RE}=
    \frac{1}{2}\xi(\frac{1}{Y-Y_{l,u}-l_w/2})^{2},
    \label{pf_road}
\end{equation}
where
\begin{equation*}
\centering
    Y_l=
    \left\{
    \begin{aligned}
    Y_{bottom}, & & {if\; X < X_{merge}^{start}} \\
    k_{sl}X_{merge}+b, & & {if\; X_{merge}^{start} \leq X \leq X_{merge}^{end}} \\
    Y_{lane}, & & {if\; X > X_{merge}^{end} }
    \end{aligned}
    \right.
\end{equation*}
with $\xi$ denoting the scaling coefficient and $Y_{l,u}$ representing the lateral positions of the road edges, respectively. $l_w$ is the vehicle's width, and $Y_{bottom}$ denotes the lateral position of the first section of the road edge. $k_{sl}$ is the slope, and $b$ is the bias, which we assume the edge of the merge section can be approximated by the linear function. $X_{merge}^{start,\;end}$ means the longitudinal start and end of the merge part.

\subsubsection{Obstacle Potential Function} The PF for an obstacle serves two main purposes. Firstly, it creates a safety zone around the obstacle by assigning a high-risk value, which helps keep the host vehicle at a safe distance. Secondly, the PF generates the path for a lane change, usually along the edge of the obstacle's PF. Therefore, the following obstacle potential function $U_{OB}$ is indicated~\cite{Lin2022-ge}:
\begin{align}
    U_{OB}=
    A_{obs} \exp{[-\frac{C_1}{2}(\frac{(X-X_o)^2}{\sigma_x}+\frac{(Y-Y_o)^2}{\sigma_y}-C_2)]},
    \label{pf_obs}
\end{align}
where
\begin{align*}
    &C_1=1-\psi_o^2,\quad
    C_2=\frac{2\psi_o(X-X_o)(Y-Y_o)}{\sigma_x\sigma_y},\\
    &\sigma_x=D_{min}\sqrt{-\frac{1}{\ln{U}}},\quad
    \sigma_y=\sqrt{-\frac{(Y-Y_o)^2}{2\ln{\frac{\epsilon}{A_{obs}}}}},\\
    &D_{min}=\frac{Mv^2}{2a_b}-\frac{M_ov_o^2}{2a_{b,o}}+\frac{l_{fr}+l_{fr,o}}{2}.
\end{align*}
with $X_o$ and $Y_o$ denoting the longitudinal and lateral position of the obstacle vehicle, respectively. $\psi_o$ means the heading angle of the obstacle vehicle, and $\epsilon$ is a minimum positive factor. $M$ and $M_o$ refer to the weights of the host and obstacle vehicles, respectively. $v$ and $v_o$ denote the longitudinal speeds of the host and obstacle vehicles, respectively. $a_b$ and $a_{b,o}$ are the maximum braking deceleration of the host and obstacle vehicles, respectively. $l_{fr}$ and $l_{fr,o}$ represent the wheelbase of the host and obstacle vehicles, respectively.

\subsubsection{Universal Potential Field} With Eqs. (\ref{attr_pf})-(\ref{pf_obs}), we can get the universal PF $U_{uni}$ by summarizing them and then applying the gradient descent method to obtain the virtual force $F_{uni}$. 
\begin{equation}
    F_{uni}=-\nabla U_{uni}=-\begin{bmatrix}\frac{\partial U_{uni}}{\partial X}  &  \frac{\partial U_{uni}}{\partial Y}\end{bmatrix}^T,
\end{equation}
where
\begin{equation*}
    U_{uni}=
    U_{AT}+U_{LD}+U_{RE}+\sum U_{OB}^j,
\end{equation*}
with $j$ denotes the $j^{th}$ obstacle. Then, with the obtained virtual force $F_{uni}$, we can calculate the AV's desired heading angle $\psi_{ref}$ for the next time step.
\begin{equation}
    \psi_{ref}=\arctan (\frac{F_{uni}(Y)}{F_{uni}(X)}).
\end{equation}
Finally, a queue of local waypoints $(X_{pf},\;Y_{pf})$ is acquired by defining a step length $l_s$ with the given $\psi_{ref}$.
\begin{align}
    \begin{cases}
        X_{pf}=X+L_s cos(\psi_{ref}) \\
        Y_{pf}=Y+L_s sin(\psi_{ref})
    \end{cases}.
    \label{pre_waypoints}
\end{align}
Therefore, the PF modeling of the merging road is depicted in Fig. \ref{inte_apf}. As mentioned in Sec. \ref{related_work}, previous researches assume that only the ego vehicle has the PF-based path planner while other AVs drive along a preset path. While the path generation of the ego vehicle may meet expectations, the impact of the ego vehicle on other AVs is often ignored or not discussed. We have discovered that if other AVs also utilize a PF-based path planner, the PF of the ego vehicle can affect the path generation of other vehicles. This consideration is particularly important in merging scenarios where cooperation is usually required from other road users.

\subsubsection{Interactive Speed Optimization} To consider the impact of other AVs' PF, we propose an interactive speed optimization with the shared local path via V2V communication for cooperative driving. First, we introduce the cubic polynomial to handle the discrete waypoints, resulting in a smooth and easy-to-follow path \cite{Nagy2001-nv, Jeon2015-vd}, as denoted below.
\begin{equation}
    f(x) = a_0 + a_1x + a_2x^2 + a_3x^3,
    \label{cubic}
\end{equation}
where $a_0,\;a_1,\;a_2,\;a_3$ are the parameters of the cubic polynomial and $x$ is the waypoints. Next, we should reshape the Eq. (\ref{cubic}) to be a quadratic programming (QP) form for the optimization with vehicle dynamics constraints:
\begin{alignat}{2}
    \mathbf{a}^* \, = \, &
    \arg \min_{\mathbf{a}}\frac{1}{2}\left(\textbf{X}\mathbf{a}-\textbf{Y}\right)^TW\left(\textbf{X}\mathbf{a}-\textbf{Y}\right)\\
    \mathrm{s.t.} \quad & \textbf{a}_{min}\preceq \textbf{a}\preceq \textbf{a}_{max}\tag{9a}
\label{cost}
\end{alignat}
where
\begin{align*}
    &\textbf{\textrm{X}}=\begin{bmatrix} 1& x_1& {x_1}^2& {x_1}^3\\ 1& x_2& {x_2}^2& {x_2}^3\\ \vdots& \vdots& \vdots& \vdots\\ 1& x_N& {x_N}^2& {x_N}^3 \end{bmatrix},
    W=\begin{bmatrix}
    w_1 & 0  & \cdots   & 0  \\
    0 & w_2  & \cdots   & 0  \\
    \vdots & \vdots  & \ddots   & \vdots  \\
    0 & 0  & \cdots\  & w_N \end{bmatrix},\\
    &\mathbf{a}=\begin{bmatrix} a_0& a_1& a_2& a_3 \end{bmatrix}^T,~
    \textbf{\textrm{Y}}=\begin{bmatrix} y_1& y_2& \cdots& y_N \end{bmatrix}^T.
    \label{waypoint_matrix}
\end{align*}
where $\mathbf{a}$ denotes the optimal cubic coefficients and $(x_N,\;y_N)$ represents the $N^{th}$ waypoint from the PF. $w_N$ is the weight. Note that at least 4 waypoints are needed for calculating $\textbf{a}$. $\textbf{a}_{min,\;max}$ is explained in \cite{Lin2022-jm}. After processing the local path according to Eq. (9), the next step is to compute the speed. To achieve this, we have formulated an objective function with reasonable constraints to obtain the optimal speed. The proposed objective function consists of three terms, which are shown below:
\begin{alignat}{2} 
\centering
    \min_{v_{1},\cdots,v_{h}} w_{1}\sum_{i=1}^{N}\frac{X_{pf}^{i+1}-X_{pf}^i}{v_i}+
    w_{2}&\sum_{i=1}^{N}\frac{1}{2}(v_i-V_{target})^2\\ \nonumber
    &+w_{3}\sum_{i=1}^{N}\ln U_{SLP}^{i}\\
    \label{object_fun}
    s.t.\ v_{i}\in[0,\ V_{i}^{max}],\tag{10a} 
\end{alignat}
where
\begin{align*}
\centering
    &V_{i}^{max}=
    \min(V_{limit},\;\sqrt{\mu g/\kappa}),\\
    &\kappa=6a_3s+2a_2\approx 2a_2
\end{align*}
with $V_{target}$ denoting the preset target speed and $V_{limit}$ indicating the speed limit on the highway. $\mu$ is the friction coefficient and $g$ is the gravitational acceleration. $\kappa$ is the path curvature, and $s$ is the arc length. $U_{SLP}$ represents the accumulative PF of the shared local path (SLP) from other AVs via V2V communication. The calculation of $U_{SLP}$ is slightly different from Eq. (\ref{pf_obs}) that we use $(X_{pf}^{ego,i},\;Y_{pf}^{ego,i})$ and $(X_{pf}^{obs,i},\;Y_{pf}^{obs,i})$ from Eq. (\ref{pre_waypoints}) instead of $(X,\;Y)$ and $(X_o,\;Y_o)$. Note that the number $i=1,\cdots,N$ of the SLP depends on the preset number of iterations. Additionally, the first term of the objective function (10) evaluates the driving efficiency while tracking the local waypoints. The second term indicates the deviation from the target (reference) speed, while the last term ensures that the AV is aware of the intentions of surrounding vehicles by computing the PF of the SLP. Therefore, we can diminish the oscillations caused by the interactive effect of other AVs' PFs. Regarding the communication network, we have designed the maneuver coordination (MC) protocol \cite{Mizutani2021-ol} and established a systematic message broadcasting module to facilitate the exchange of anticipated paths among the vehicles \cite{Hirata2021-dr} in our previous works. Therefore, in this study, we utilize the previous communication network, concentrating more on path generation and speed computation for connected AVs.

\section{Simulation Results}\label{AA}
\begin{figure*}[ht]
\centering
    \includegraphics[width=\hsize]{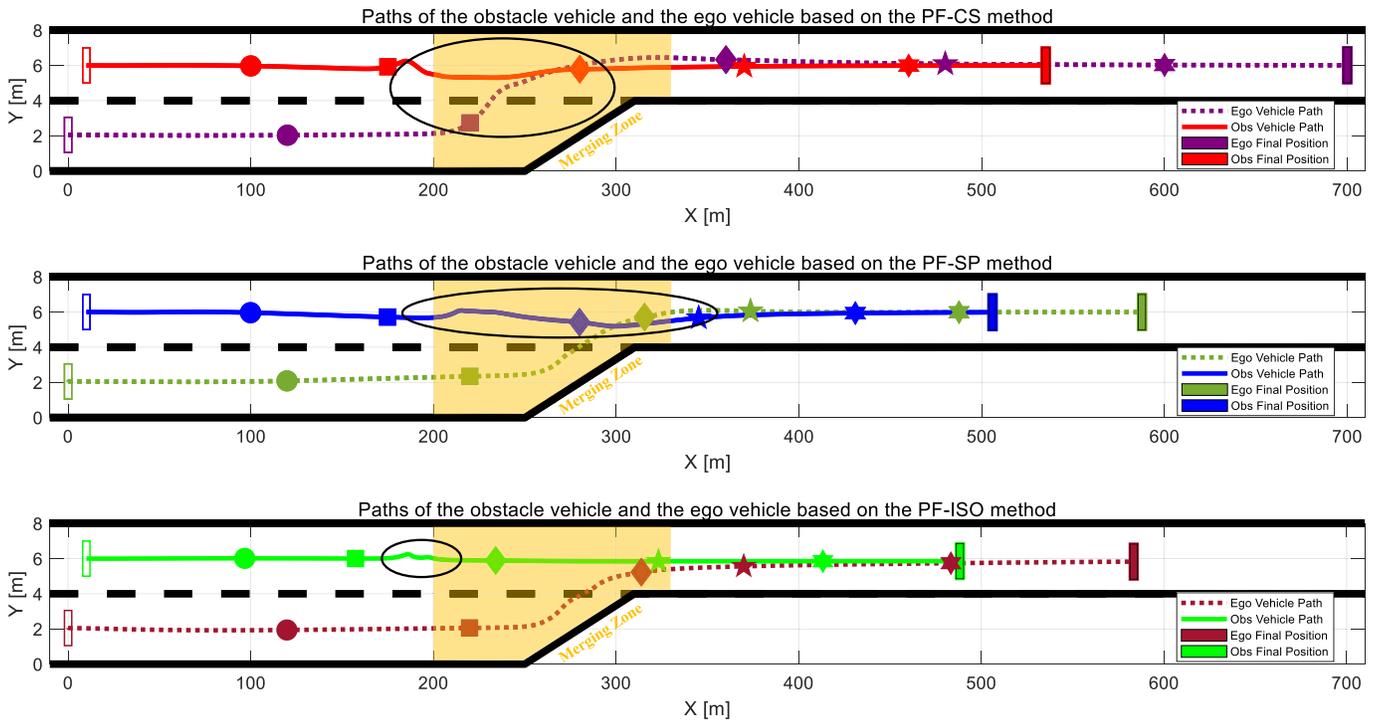}
    \caption{Paths of the obstacle and ego vehicles based on PF-CS, PF-SP, and PF-ISO, respectively}
    \label{paths}
\end{figure*}

In this section, we compared our proposed method with other PF-based path-planning algorithms in a merging scenario using MATLAB/Simulink. To simulate interaction, we assumed that all AVs utilized the PF for local path planning. As shown in Fig. \ref{paths}, there are two autonomous vehicles on a merging road. An obstacle vehicle drives at an initial speed of 15 m/s in the upper lane, while the ego vehicle with an initial speed of 20 m/s drives in the lower lane and needs to perform a merging maneuver. Note that we intentionally positioned the obstacle vehicle slightly ahead of the ego vehicle to simulate an urgent merging situation, which provides better observation of the paths of both vehicles. In addition, we denoted the following path planners for comparative study: \romannumeral1) PF-based planner with longitudinally constant speed (denoted as PF-CS) \cite{Wang2022-tt}; \romannumeral2) PF-based planner with speed planning (denoted as PF-SP) \cite{Wu2022-ae}; \romannumeral2) Proposed PF-based planner with interactive speed optimization (denoted as PF-ISO).

\begin{figure*}
\centering
    \subfigure[Sideslip angle]{
        \begin{minipage}[t]{0.5\linewidth}
            \centering
            \includegraphics[width=\hsize]{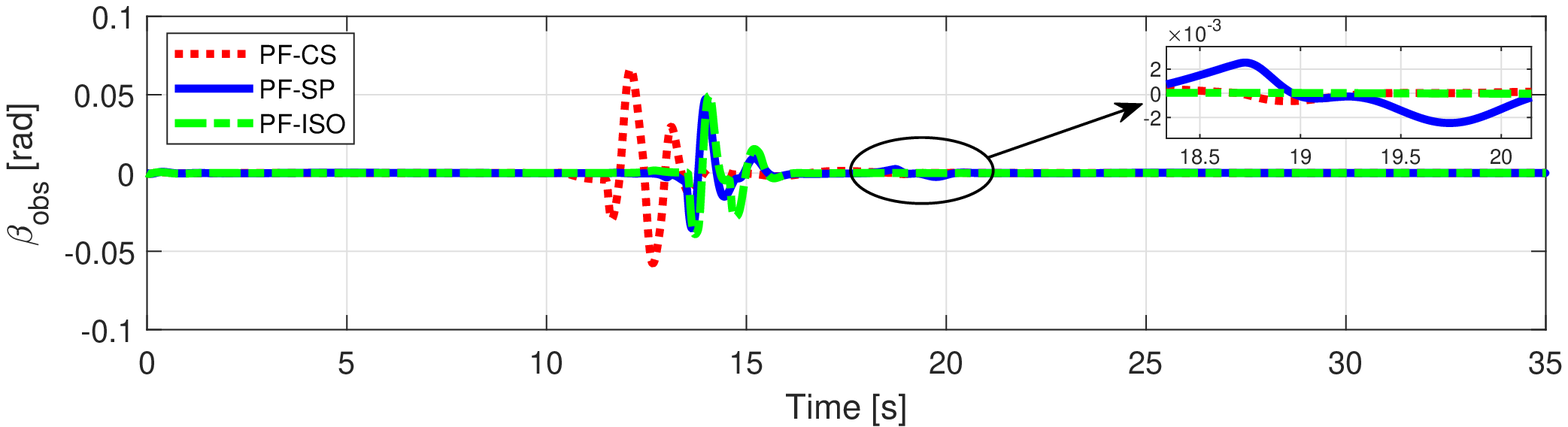}
        \end{minipage}
    \label{beta}
    }%
    \subfigure[Yaw angle]{
        \begin{minipage}[t]{0.5\linewidth}
            \centering
            \includegraphics[width=\hsize]{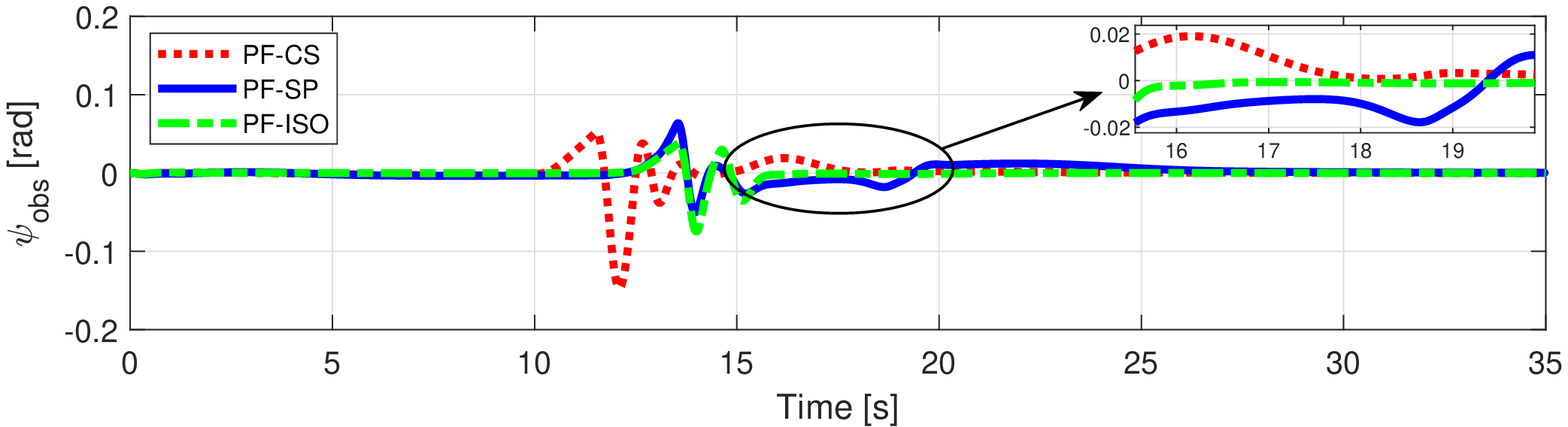}
        \end{minipage}
    \label{psi}
    }

    \subfigure[Yaw rate]{
        \begin{minipage}[t]{0.5\linewidth}
            \centering
            \includegraphics[width=\hsize]{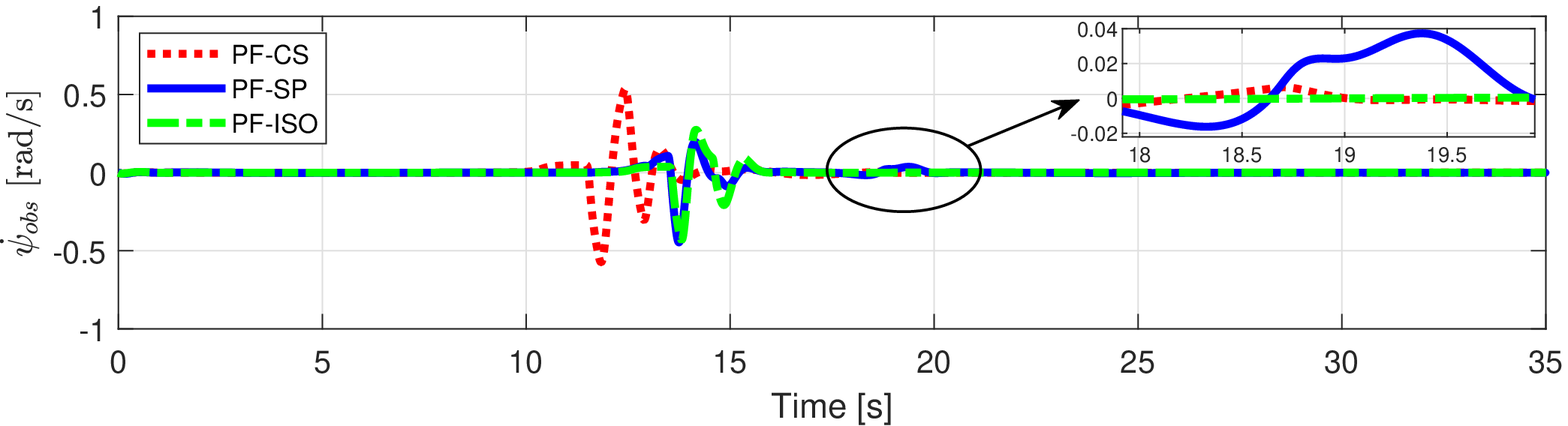}
        \end{minipage}
    \label{psi_dot}
    }%
    \subfigure[Longitudinal speed]{
        \begin{minipage}[t]{0.5\linewidth}
            \centering
            \includegraphics[width=\hsize]{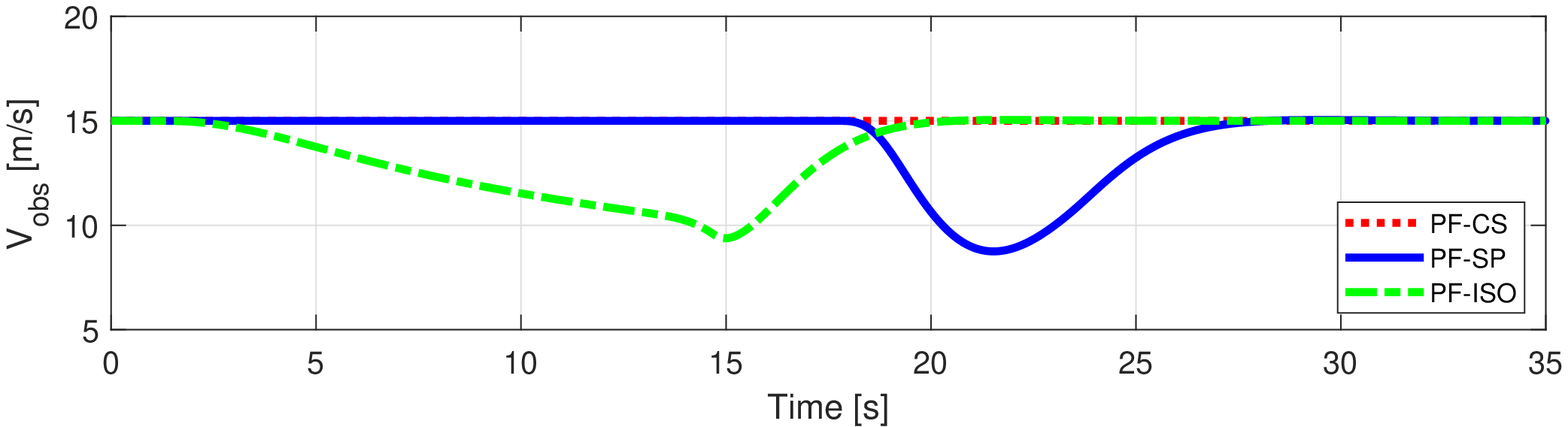}
        \end{minipage}
    \label{long_vel}
    }
\caption{Motion states of the obstacle vehicle based on PF-CS, PF-SP, and PF-ISO, respectively}
\end{figure*}

In Fig. \ref{paths}, we can observe the generated paths of both the obstacle and ego vehicles from the three PF-based path planners. In the first subfigure of Fig. \ref{paths}, we can see that the red path (obstacle vehicle) has a winding part from 167 m to 272 m with the PF-CS method, as circled in the ellipse when entering the merging zone (orange area). The winding part can also be observed in the ego vehicle's path (brown-purple dash line). Hence, we can see that the sideslip angle, yaw angle, and yaw rate of the PF-CS planner have larger values than the other two planners, shown as the red dash line in Figs. \ref{beta} to \ref{psi_dot} (with a maximum of 0.07 rad in sideslip angle, -0.15$^\circ$ in yaw angle, and -0.57 rad/s in yaw rate). Conversely, the winding part is slightly decreased with the PF-SP planner for both the obstacle and ego vehicles' paths, as depicted in the second subfigure of Fig. \ref{paths} (blue solid and green dashed lines). However, it is still obvious that the obstacle vehicle has a serpentine part after the ego vehicle finishes the merging maneuver from 279 m to 345 m. Therefore, we can see a certain vibration in the sideslip angle, yaw angle, and yaw rate (blue solid line) from 18 s to 20 s in Figs. \ref{beta} to \ref{psi_dot}. In Fig. \ref{long_vel}, we can also see the PF-SP planner reduces the speed by around 17.5 s which corresponds to the merging part at 255 m in the second subfigure of Fig. \ref{paths}. In contrast, the PF-ISO planner starts to slow down gradually from 3.1 s to 14.9 s (from 15 m/s to 9.4 m/s) in Fig. \ref{long_vel} and then accelerates until reaching the target longitudinal speed. Furthermore, the motion states of the PF-ISO planner vary around zero, which indicates it is less affected by the interaction between the PFs during the merging procedure, as denoted in the partial magnifications of Figs. \ref{beta}, \ref{psi}, and \ref{psi_dot}. In addition, we can still observe a slight oscillation in the obstacle vehicle's path (solid green line) from 180 m to 196 m in the third subfigure of Fig. \ref{paths}, which implicitly states the ego vehicle's PF continuously influences the obstacle vehicle's path generation. However, the influence is significantly reduced because the obstacle vehicle generally decelerates before closing to the merging part. It is worth noting that the path length of the PF-ISO is relatively shorter than these of the other two planners, which implies the PF-ISO method performs a pre-brake maneuver like human drivers.

%
%
\vspace{-0.3cm}
\section{Conclusion}

In this paper, we have introduced a unique PF-based path planning approach with the SLP path from other AVs via V2V communication. We consider the interaction between the PFs of the AVs on the road instead of assuming that only the ego AV uses the PF-based path planner. By formulating an objective function, we compute the optimal speed that implements the SLP as an optimized term, which also imposes reasonable constraints. The simulation results indicate that the proposed PF-ISO method can reduce the oscillations of the AVs' sideslip and heading angles and produce a smoother path for both the obstacle and the ego vehicles under the interaction of other AVs' PFs. In future work, we will dig deeper into the driving uncertainty of road users, including pedestrians, bikes, etc. Besides, various driving scenarios will be applied for comprehensive evaluations, as well as real vehicle experiments.

\section*{Acknowledgment}

These research results were obtained from the commissioned research Grant number 01101 by the National Institute of Information and Communications Technology (NICT), Japan, and the Japan Society for the Promotion of Science (JSPS) KAKENHI (grant number: 21H03423), and partly sponsored by the China Scholarship Council (CSC) program (No.202208050036) and JSPS DC program (grant number: 23KJ0391).

\bibliographystyle{IEEEtran}
\bibliography{Reference.bib}

\begin{thebibliography}{10}
\providecommand{\url}[1]{#1}
\csname url@samestyle\endcsname
\providecommand{\newblock}{\relax}
\providecommand{\bibinfo}[2]{#2}
\providecommand{\BIBentrySTDinterwordspacing}{\spaceskip=0pt\relax}
\providecommand{\BIBentryALTinterwordstretchfactor}{4}
\providecommand{\BIBentryALTinterwordspacing}{\spaceskip=\fontdimen2\font plus
\BIBentryALTinterwordstretchfactor\fontdimen3\font minus
  \fontdimen4\font\relax}
\providecommand{\BIBforeignlanguage}[2]{{%
\expandafter\ifx\csname l@#1\endcsname\relax
\typeout{** WARNING: IEEEtran.bst: No hyphenation pattern has been}%
\typeout{** loaded for the language `#1'. Using the pattern for}%
\typeout{** the default language instead.}%
\else
\language=\csname l@#1\endcsname
\fi
#2}}
\providecommand{\BIBdecl}{\relax}
\BIBdecl

\bibitem{AVreport2021}
D.~of~Motor~Vehicles, ``\BIBforeignlanguage{en}{Autonomous vehicle collision
  reports},'' \emph{\BIBforeignlanguage{en}{State of California}}, Oct. 2021,
  \url{https://www.dmv.ca.gov/portal/vehicle-industry-services/autonomous-vehicles/autonomous-vehicle-collision-reports/}.

\bibitem{McCarthy2022-fp}
R.~L. McCarthy, ``Autonomous vehicle accident data analysis: California {OL}
  316 reports: 2015--2020,'' \emph{ASCE-ASME J Risk and Uncert in Engrg Sys
  Part B Mech Engrg}, vol.~8, no.~3, Sep. 2022.

\bibitem{Khatib1986-dv}
O.~Khatib, ``Real-time obstacle avoidance for manipulators and mobile robots,''
  in \emph{Autonomous Robot Vehicles}.\hskip 1em plus 0.5em minus 0.4em\relax
  New York, NY: Springer New York, 1986, pp. 396--404.

\bibitem{Lopez1995-pr}
I.~Lopez and C.~R. Mclnnes, ``Autonomous rendezvous using artificial potential
  function guidance,'' \emph{Journal of guidance, control, and dynamics: a
  publication of the American Institute of Aeronautics and Astronautics devoted
  to the technology of dynamics and control}, vol.~18, no.~2, pp. 237--241,
  Mar. 1995.

\bibitem{Leonard2001-ir}
N.~E. Leonard and E.~Fiorelli, ``Virtual leaders, artificial potentials and
  coordinated control of groups,'' in \emph{Proceedings of the 40th {IEEE}
  Conference on Decision and Control (Cat. {No.01CH37228})}, vol.~3, Dec. 2001,
  pp. 2968--2973 vol.3.

\bibitem{Baronov2008-zf}
D.~Baronov and J.~Baillieul, ``Autonomous vehicle control for
  ascending/descending along a potential field with two applications,'' in
  \emph{2008 American Control Conference}, Jun. 2008, pp. 678--683.

\bibitem{Wolf2008-ye}
M.~T. Wolf and J.~W. Burdick, ``Artificial potential functions for highway
  driving with collision avoidance,'' in \emph{2008 {IEEE} International
  Conference on Robotics and Automation}, May 2008, pp. 3731--3736.

\bibitem{Koren1991-ar}
Y.~Koren and J.~Borenstein, ``Potential field methods and their inherent
  limitations for mobile robot navigation,'' in \emph{Proceedings. 1991 {IEEE}
  International Conference on Robotics and Automation}, Apr. 1991, pp.
  1398--1404 vol.2.

\bibitem{Hesse2007-ho}
T.~Hesse and T.~Sattel, ``An approach to integrate vehicle dynamics in motion
  planning for advanced driver assistance systems,'' in \emph{2007 {IEEE}
  Intelligent Vehicles Symposium}, Jun. 2007, pp. 1240--1245.

\bibitem{Ji2017-fk}
J.~Ji, A.~Khajepour, W.~W. Melek, and Y.~Huang, ``Path planning and tracking
  for vehicle collision avoidance based on model predictive control with
  multiconstraints,'' \emph{IEEE Transactions on Vehicular Technology},
  vol.~66, no.~2, pp. 952--964, Feb. 2017.

\bibitem{Wang2019-nf}
H.~Wang, Y.~Huang, A.~Khajepour, Y.~Zhang, Y.~Rasekhipour, and D.~Cao, ``Crash
  mitigation in motion planning for autonomous vehicles,'' \emph{IEEE
  Transactions on Intelligent Transportation Systems}, vol.~20, no.~9, pp.
  3313--3323, Sep. 2019.

\bibitem{Wang2020-xl}
H.~Wang, Y.~Huang, A.~Khajepour, D.~Cao, and C.~Lv, ``Ethical {Decision-Making}
  platform in autonomous vehicles with lexicographic optimization based model
  predictive controller,'' \emph{IEEE Transactions on Vehicular Technology},
  vol.~69, no.~8, pp. 8164--8175, Aug. 2020.

\bibitem{Lin2020-ry}
P.~Lin, W.~Y. Choi, and C.~C. Chung, ``Local path planning using artificial
  potential field for waypoint tracking with collision avoidance,'' in
  \emph{2020 {IEEE} 23rd International Conference on Intelligent Transportation
  Systems ({ITSC})}.\hskip 1em plus 0.5em minus 0.4em\relax
  ieeexplore.ieee.org, Sep. 2020, pp. 1--7.

\bibitem{Lin2020-uc}
P.~Lin, W.~Y. Choi, S.-H. Lee, and C.~C. Chung, ``Model predictive path
  planning based on artificial potential field and its application to
  autonomous lane change,'' in \emph{2020 20th International Conference on
  Control, Automation and Systems ({ICCAS})}, Oct. 2020, pp. 731--736.

\bibitem{Xie2022-jm}
S.~Xie, J.~Hu, P.~Bhowmick, Z.~Ding, and F.~Arvin, ``Distributed motion
  planning for safe autonomous vehicle overtaking via artificial potential
  field,'' \emph{IEEE Transactions on Intelligent Transportation Systems}, pp.
  1--17, 2022.

\bibitem{Wang2022-tt}
J.~Wang, Y.~Yan, K.~Zhang, Y.~Chen, M.~Cao, and G.~Yin, ``Path planning on
  large curvature roads using {Driver-Vehicle-Road} system based on the
  kinematic vehicle model,'' \emph{IEEE Transactions on Vehicular Technology},
  vol.~71, no.~1, pp. 311--325, Jan. 2022.

\bibitem{Wu2022-ae}
P.~Wu, F.~Gao, and K.~Li, ``Humanlike decision and motion planning for
  expressway lane changing based on artificial potential field,'' \emph{IEEE
  Access}, vol.~10, pp. 4359--4373, 2022.

\bibitem{Lin2022-ge}
P.~Lin and M.~Tsukada, ``Adaptive potential field with collision avoidance for
  connected autonomous vehicles,'' in \emph{2022 13th Asian Control Conference
  ({ASCC})}, May 2022, pp. 2251--2256.

\bibitem{Nagy2001-nv}
B.~Nagy and A.~Kelly, ``Trajectory generation for car-like robots using cubic
  curvature polynomials,'' \emph{Field and Service Robots}, vol.~11, pp.
  479--490, 2001.

\bibitem{Jeon2015-vd}
S.~J. Jeon, C.~M. Kang, S.-H. Lee, and C.~C. Chung, ``{GPS} waypoint fitting
  and tracking using model predictive control,'' in \emph{2015 {IEEE}
  Intelligent Vehicles Symposium ({IV})}, Jun. 2015, pp. 298--303.

\bibitem{Lin2022-jm}
P.~Lin, J.~H. Yang, Y.~S. Quan, and C.~C. Chung,
  ``\BIBforeignlanguage{en}{Potential field‐based path planning for emergency
  collision avoidance with a clothoid curve in waypoint tracking},''
  \emph{\BIBforeignlanguage{en}{Asian journal of control}}, vol.~24, no.~3, pp.
  1074--1087, May 2022.

\bibitem{Mizutani2021-ol}
M.~Mizutani, M.~Tsukada, and H.~Esaki, ``{AutoMCM}: Maneuver coordination
  service with abstracted functions for autonomous driving,'' in \emph{2021
  {IEEE} International Intelligent Transportation Systems Conference ({ITSC})},
  Sep. 2021, pp. 1069--1076.

\bibitem{Hirata2021-dr}
M.~Hirata, M.~Tsukada, K.~Okumura, Y.~Tamura, H.~Ochiai, and X.~D{\'e}fago,
  ``{Roadside-Assisted} cooperative planning using future path sharing for
  autonomous driving,'' in \emph{2021 {IEEE} 94th Vehicular Technology
  Conference ({VTC2021-Fall})}, Sep. 2021, pp. 1--7.

\end{thebibliography}

\end{document}